\documentclass[10pt,twocolumn,letterpaper]{article}

\usepackage{iccv}
\usepackage{times}
\usepackage{epsfig}
\usepackage{graphicx}
\usepackage{amsmath}
\usepackage{amssymb}

\usepackage{booktabs}
\usepackage{multirow}
\usepackage{siunitx}
\usepackage{subcaption}
\usepackage{xcolor,colortbl}

\definecolor{mygray}{gray}{0.93}

\usepackage[breaklinks=true, bookmarks=false, colorlinks=true]{hyperref}

\iccvfinalcopy 

\def\iccvPaperID{8833} 

\ificcvfinal\fi

\begin{document}

\title{Instances as Queries}

\author{Yuxin Fang$^{1*}$, \ \ Shusheng Yang$^{1,2}$\thanks{Equal contributions. This work was done while Shusheng Yang was interning at Applied Research Center (ARC), Tencent PCG.}, \ \  Xinggang Wang$^{1}$\thanks{Corresponding author, E-mail: {\tt xgwang@hust.edu.cn.}}, \ \ Yu Li$^{2}$, \\
\vspace{0.15cm}
Chen Fang$^{3}$, \ \ Ying Shan$^{2}$, \ \ Bin Feng$^{1}$, \ \ Wenyu Liu$^{1}$ \\

$^1$School of EIC, Huazhong University of Science \& Technology \\
$^2$Applied Research Center (ARC), Tencent PCG \ \ $^3$Tencent\\
}

\maketitle
\ificcvfinal\thispagestyle{empty}\fi

\begin{abstract}
Recently, query based object detection frameworks achieve comparable performance with previous state-of-the-art object detectors. However, how to fully leverage such frameworks to perform instance segmentation remains an open problem. In this paper, we present QueryInst (Instances as Queries), a query based instance segmentation method driven by parallel supervision on dynamic mask heads. The key insight of QueryInst is to leverage the intrinsic one-to-one correspondence in object queries across different stages, as well as one-to-one correspondence between mask RoI features and object queries in the same stage. This approach eliminates the explicit multi-stage mask head connection and the proposal distribution inconsistency issues inherent in non-query based multi-stage instance segmentation methods. We conduct extensive experiments on three challenging benchmarks, \ie, COCO, CityScapes, and YouTube-VIS to evaluate the effectiveness of QueryInst in instance segmentation and video instance segmentation (VIS) task. Specifically, using ResNet-101-FPN backbone, QueryInst obtains 48.1 box AP and 42.8 mask AP on COCO test-dev, which is 2 points higher than HTC in terms of both box AP and mask AP, while runs 2.4 times faster. For video instance segmentation, QueryInst achieves the best performance among all online VIS approaches and strikes a decent speed-accuracy trade-off. Code is available at \url{https://github.com/hustvl/QueryInst}.
\end{abstract}

\section{Introduction}


Instance segmentation is a fundamental yet challenging computer vision task that requires an algorithm to assign a pixel-level mask with a category label for each instance of interest in image.
Prevalent state-of-the-art instance segmentation methods are based on high performing object detectors and follow a multi-stage paradigm.
Among which, the Mask R-CNN family~\cite{MaskRCNN, MaskScoring, PANet, CascadeRCNN_TPAMI, HTC, SparseRCNN} is the most successful one, where the regions-of-interest (RoI) for instance segmentation is extracted via a region-wise pooling operation (\eg, RoIPool~\cite{SPP, FastRCNN} or RoIAlign~\cite{MaskRCNN}) based on the box-level localization information from the region proposal network (RPN)~\cite{FasterRCNN}, or the previous stage bounding-box prediction~\cite{CascadeRCNN, CascadeRCNN_TPAMI}. 
The final instance mask is obtained via feeding the RoI feature into the mask head, which is a small fully convolutional network (FCN)~\cite{FCN}.


\begin{figure}[t]
  \centering
  \includegraphics[width=0.9\columnwidth]{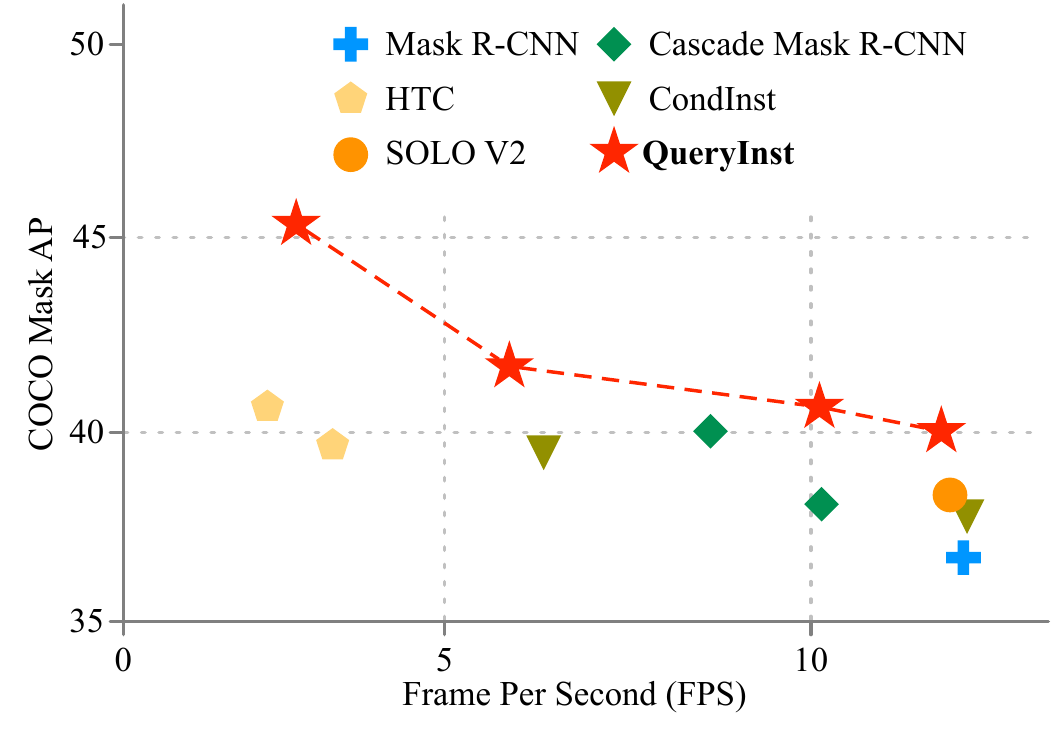}
  \caption{AP \vs FPS on COCO $\mathtt{test}$-$\mathtt{dev}$. 
  QueryInst outperforms current state-of-the-art methods in terms of both accuracy and speed.
  The speed is measured using a single Titan Xp GPU.}
  \label{fig: teaser}
\end{figure}

Recently, DETR~\cite{DETR} is proposed to reformulate object detection as a query based direct set prediction problem, whose input is mere $100$ learned object queries. 
Follow-up works~\cite{DefDETR, SparseRCNN, TSP, SMCA, ACT, UP-DETR} in object detection improve this query based approach and achieve comparable performance with state-of-the-art detectors such as Cascade R-CNN~\cite{CascadeRCNN}. 
The results show that query based instance-level perception is a very promising research direction.
Thus, enabling query based detection framework to perform instance segmentation is highly desirable. 
However, we find that it is inefficient to integrate the previous successful practices in Cascade Mask R-CNN~\cite{CascadeRCNN_TPAMI} and HTC~\cite{HTC}, which are state-of-the-art mask generation solutions in the non-query based paradigm, directly into query based detectors for instance mask generation.
Therefore, an instance segmentation method tailored for the query based end-to-end framework is urgently needed.

To bridge this gap, we propose \textbf{QueryInst} (Instances as Queries), a query based end-to-end instance segmentation method driven by parallel supervision on dynamic mask heads~\cite{DFN, CondInst, SparseRCNN}.
The key insight of QueryInst is to leverage \textit{the intrinsic one-to-one correspondence in object queries} across different stages, and \textit{one-to-one correspondence between mask RoI features and object queries} in the same stage.
Specifically, we set up dynamic mask heads in parallel with each other, which transform each mask RoI feature adaptively according to the corresponding query, and are simultaneously trained in all stages.
The mask gradient not only flows back to the backbone feature extractor, but also to the object query, which is intrinsically one-to-one interlinked in different stages.
The queries implicitly carry the multi-stage mask information, which is read by RoI features in dynamic mask heads for final mask generation.
There is no explicit connection between different stage mask heads or mask features.
Moreover, the queries are shared between object detection and instance segmentation sub-networks in each stage, enabling cross-task communications that one task can take advantage of the information from the other task.
We demonstrate that this shared query design can fully leverage the synergy between object detection and instance segmentation.
When the training is completed, we throw away all the dynamic mask heads in the intermediate stages and only use the final stage predictions for inference.
Under such a scheme, QueryInst surpasses the state-of-the-art HTC in terms of AP while runs much faster.
Concretely, our main contributions are summarized as follows:
\begin{itemize}

    \item We attempt to solve instance segmentation from a new perspective that uses parallel dynamic mask heads in the query based end-to-end detection framework. 
    This novel solution enables such a new framework to outperform well-established and highly-optimized non-query based multi-stage schemes such as Cascade Mask R-CNN and HTC in terms of both accuracy and speed (see Fig.~\ref{fig: teaser}).
    Specifically, using ResNet-$101$-FPN backbone~\cite{ResNet, FPN}, QueryInst obtains $48.1$ AP$^{\mathtt{box}}$ and $42.8$ AP$^{\mathtt{mask}}$ on COCO $\mathtt{test}$-$\mathtt{dev}$, which is $2$ point higher than HTC in terms of both box AP and mask AP, while runs $2.4 \times$ faster.
    Without bells and whistles, our best model achieves $50.4$ AP$^{\mathtt{box}}$ and $46.6$ AP$^{\mathtt{mask}}$ on COCO $\mathtt{test}$-$\mathtt{dev}$.
    
    \item We set up a task-joint paradigm for query based object detection and instance segmentation by leveraging the shared query and multi-head self-attention design. 
    This paradigm establishes a kind of communication and synergy between detection and segmentation tasks, which encourages this two tasks to benefits from each other.
    We demonstrate that our architecture design can also significantly improve the object detection performance.

    
    
	\item We extend the QueryInst to video instance segmentation task (VIS)~\cite{VIS} task by simply adding a vanilla track head. 
	Experiments on YouTube-VIS dataset~\cite{VIS} indicate that with same tracking approach, our methods outperforms MaskTrack R-CNN~\cite{VIS} and SipMask-VIS~\cite{SipMask} by a large margin. 
	QueryInst-VIS can even outperform well-designed VIS approaches such as STEm-Seg~\cite{STEm-Seg} and VisTR~\cite{VisTR}.
\end{itemize}

\section{Related Work}

\paragraph{Query Based Methods.}
Recently, query based methods emerged to tackle the set-prediction problems. 
Concretely, DETR~\cite{DETR} first introduces the query based methods with transformer architecture to object detection.
Deformable DETR~\cite{DefDETR}, UP-DETR~\cite{UP-DETR}, ACT~\cite{ACT} and TSP~\cite{TSP} improve the performance on the top of DETR.
The recently proposed Sparse R-CNN~\cite{SparseRCNN} builds a query based set-prediction framework upon R-CNN~\cite{RCNN,FastRCNN,FasterRCNN} based detector. 
For segmentation, VisTR~\cite{VisTR} introduces a query based sequence matching and segmentation method to video instance segmentation, building a fully end-to-end framework for instance segmentation in video. Max-DeepLab\cite{MaxDeepLab} presents the first box-free end-to-end panoptic segmentation model with a global memory as external query.
Trackformer~\cite{Trackformer} and Transtrack~\cite{Transtrack} build a query based multiple object tracktor upon DETR and Deformable DETR, respectively, and attain comparable results to the non-query based methods. 
AS-Net~\cite{AS-Net} introduces a query based set-prediction pipeline to human object interaction and obtains promising results. 
Despite query based set-prediction method is being widely used to many computer vision tasks, few efforts are conducted to build a successful query based instance segmentation framework. 
We aim to achieve this goal in this paper.

\paragraph{Object Detection.}
Object detection is a fundamental computer vision task which aims to detect visual objects with bounding boxes.
With the propose of R-CNN~\cite{RCNN}, Fast R-CNN~\cite{FastRCNN} and Faster R-CNN~\cite{FasterRCNN}, anchor based methods \cite{CascadeRCNN, LibraRCNN, GridRCNN, RetinaNet, SSD} dominate object detection for a long period. 
CenterNet~\cite{CenterNet} and FCOS~\cite{FCOSpro} establish anchor-free detectors with competitive detection performance.
Recently, with the proposed DETR~\cite{DETR}, query based set-prediction methods catch lots of attentions.
Deformable DETR~\cite{DefDETR} introduces deformable convolution~\cite{DCNv2} to the DETR framework, achieving better performance with faster training convergence. 
UP-DETR~\cite{UP-DETR} extends DETR to unsupervised scenarios.
ACT~\cite{ACT} and TSP~\cite{TSP} introduce the adaptive clustering module and a new bipartite matching method to DETR.
Sparse R-CNN~\cite{SparseRCNN} build a query based detector on top of R-CNN architecture, while OneNet~\cite{OneNet} and DeFCN~\cite{DeFCN} are end-to-end detector built upon the one-stage FCOS~\cite{FCOSpro}. 
In this work, we present a query based instance segmentation method on the top of the query based Sparse R-CNN detector.

\paragraph{Instance Segmentation.}

Instance segmentation is a fundamental yet challenging computer vision task that requires an algorithm to assign a pixel-level mask with a category label for each instance of interest in image.
Mask R-CNN~\cite{MaskRCNN} introduces a fully convolutional mask head to Faster R-CNN~\cite{FastRCNN} detector. 
Casacde Mask R-CNN~\cite{CascadeRCNN_TPAMI} simply combine the Casacde R-CNN~\cite{CascadeRCNN} with Mask R-CNN. HTC~\cite{HTC} presents interleaved execution and mask information flow and achieves state-of-the-art performance. 
In addition to R-CNN based methods, YOLACT~\cite{YOLACT, YOLACTpp}, SipMask~\cite{SipMask}, CondInst~\cite{CondInst_TPAMI} and SOLO~\cite{SOLO, SOLOv2} build one-stage instance segmentation framework on the top of one-stage framework, achieving comparable results with favorable inference speed.
Following the R-CNN based methods, we present a query based instance segmentation framework.


\section{Instances as Queries}

\begin{figure}[t]
  \centering
  \includegraphics[width=1.0\columnwidth]{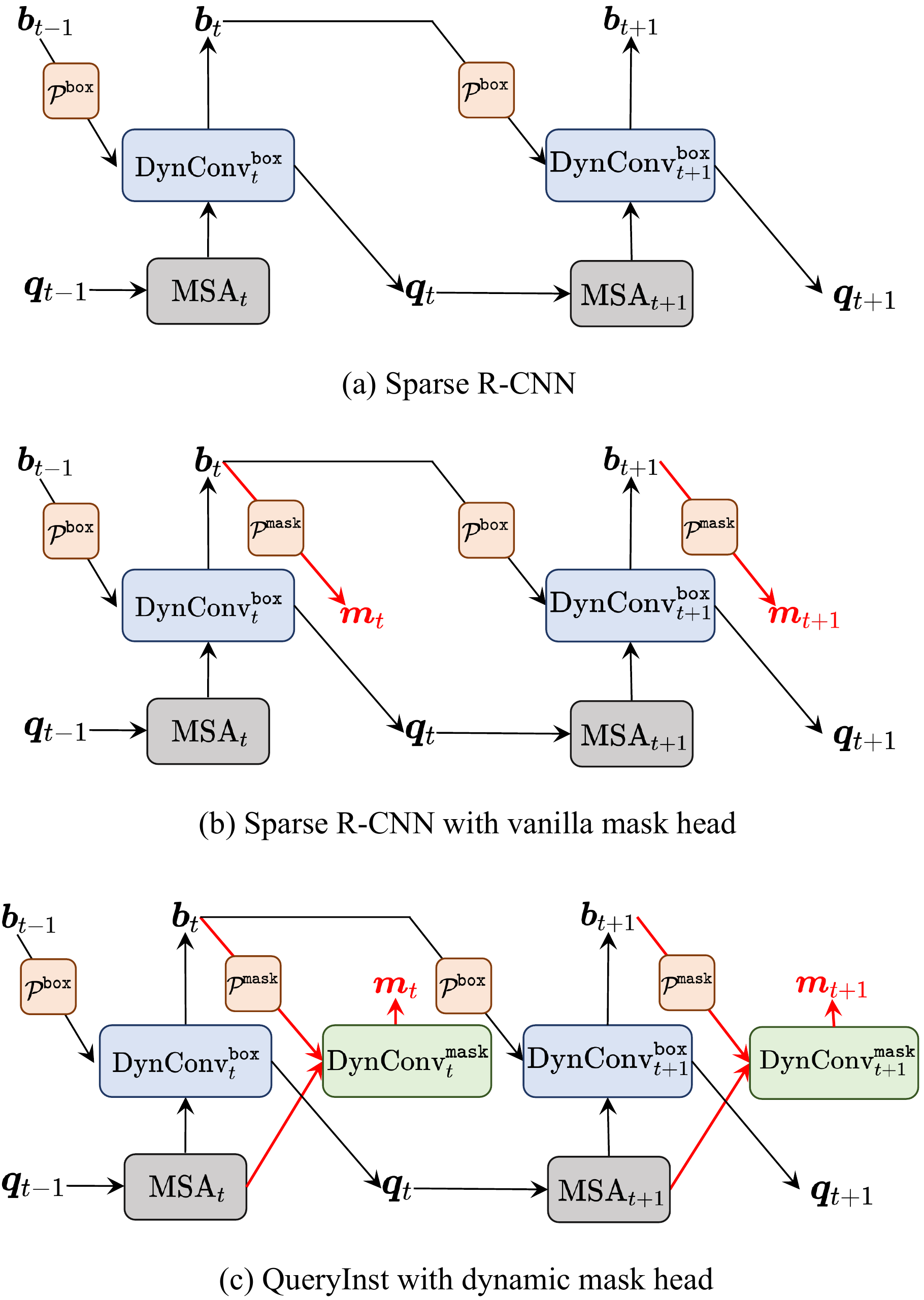}
  \caption{Overview of QueryInst. The {\color{red} red arrows} indicate mask branches. Please note that QueryInst consists of $6$ stage in parallel, \ie, $t = \{1, 2, 3, 4, 5, 6\}$. The figure only shows $2$ stages.}
  \label{fig: QueryInst}
\end{figure}

We propose QueryInst (Instances as Queries), a query based end-to-end instance segmentation method.
QueryInst consists of a query based object detector and six dynamic mask heads driven by parallel supervision.
Our key insight is to leverage the \textit{intrinsic one-to-one correspondence in queries} across different stages.
This correspondence exists in all query based framework~\cite{Transformer, BERT, ParallelWavenet, Imputer, DETR} regardless of the specific instantiations and applications.
The overall architecture of QueryInst is illustrated in Fig.~\ref{fig: QueryInst} (c).

\subsection{Query based Object Detector}
\label{sec: object detection}

QueryInst can be built on any multi-stage query based object detector~\cite{DETR, DefDETR, SparseRCNN}.
We choose Sparse R-CNN~\cite{SparseRCNN} as our default instantiation, which has six query stages.
The object detection pipeline is depicted in Fig.~\ref{fig: QueryInst} (a) and can be formulated as follows:
\begin{equation}
\begin{aligned}
    \boldsymbol{x}_{t}^{\mathtt{box}} \leftarrow & \ \mathcal{P}^{\mathtt{box}}\left(\boldsymbol{x}^{\mathtt{FPN}}, \boldsymbol{b}_{t - 1}\right), \\
    \boldsymbol{q}_{t - 1}^* \leftarrow & \ \mathrm{MSA}_{t}\left(\boldsymbol{q}_{t - 1} \right), \\
    \boldsymbol{x}_{t}^{\mathtt{box*}}, \boldsymbol{q}_{t} \leftarrow & \ \mathrm{DynConv}^{\mathtt{box}}_{t} \left(\boldsymbol{x}_{t}^{\mathtt{box}}, \boldsymbol{q}_{t - 1}^* \right), \\
    \boldsymbol{b}_{t} \leftarrow & \ \mathcal{B}_{t}\left(\boldsymbol{x}_{t}^{\mathtt{box*}}\right), \\
\end{aligned}
\end{equation}

\noindent
where $\boldsymbol{q} \in \mathbf{R}^{N \times d} $ denotes the object query. $N$ and $d$ denote the length (number) and dimension of query $\boldsymbol{q}$.
At stage $t$, a pooling operator $\mathcal{P}^{\mathtt{box}}$ extracts the current stage bounding box features $\boldsymbol{x}_{t}^{\mathtt{box}}$ from FPN~\cite{FPN} features $\boldsymbol{x}^{\mathtt{FPN}}$ under the guidance of previous stage bounding box predictions $\boldsymbol{b}_{t - 1}$.
Meanwhile, a multi-head self-attention module $\mathrm{MSA}_{t}$ is applied to the input query $\boldsymbol{q}_{t - 1}$ to get the transformed query $\boldsymbol{q}_{t - 1}^*$.
Then, a box dynamic convolution module $\mathrm{DynConv}^{\mathtt{box}}_{t}$ takes $\boldsymbol{x}_{t}^{\mathtt{box}}$ and $\boldsymbol{q}_{t - 1}^*$ as inputs and enhances the $\boldsymbol{x}_{t}^{\mathtt{box}}$ by reading $\boldsymbol{q}_{t - 1}^*$ while generating $\boldsymbol{q}_{t}$ for the next stage. 
Finally, the enhanced bounding box features $\boldsymbol{x}_{t}^{\mathtt{box}*}$ are fed into the box prediction branch $\mathcal{B}_{t}$ for current bounding box prediction $\boldsymbol{b}_{t}$.

\subsection{Mask Head Architecture}
\subsubsection{Vanilla Mask Head}
\label{sec: vanilla mask head}
For instance mask prediction, we first adopt the widely used vanilla mask head architecture design in Mask R-CNN ~\cite{MaskRCNN} as our instance segmentation baseline.
The model architecture is depicted in Fig.~\ref{fig: QueryInst} (b).
Based on the object detection pipeline described in Sec.~\ref{sec: object detection}, the mask generation process can be expressed as follows:
\begin{equation}
\begin{aligned}
    \boldsymbol{x}_{t}^{\mathtt{mask}} \leftarrow & \ \mathcal{P}^{\mathtt{mask}}\left(\boldsymbol{x}^{\mathtt{FPN}}, \boldsymbol{b}_{t}\right), \\
    \boldsymbol{m}_{t} \leftarrow & \ \mathcal{M}_{t}\left(\boldsymbol{x}_{t}^{\mathtt{mask}}\right),
\end{aligned}
\end{equation}
\noindent
where $\boldsymbol{b}_{t}$ is the bounding box predictions from the object detector. $\mathcal{P}^{\mathtt{mask}}$ denotes a region-wise pooling operator for mask RoI features extraction.
$\mathcal{M}_{t}$ indicates the mask FCN head consisting of a stack of four consecutive conv-layers, one dconv-layer and one $1 \times 1$ conv-layer for mask generation~\cite{MaskRCNN}. 
$\boldsymbol{m}_{t}$ is the current stage mask predictions. 

Overall, this vanilla design is an analogy of Cascade Mask R-CNN\cite{CascadeRCNN_TPAMI} in a query based framework.
However, we find that this design is not as effective as the original Cascade Mask R-CNN.
Moreover, establishing explicit mask flow following HTC~\cite{HTC} on top of this design (Fig.~\ref{fig: QueryInst} (b)) can only bring moderate improvements at a cost of large drops in both training and inference speed.
Part of the reasons may be the number of queries in our framework is much smaller than the number of proposals in Cascade Mask R-CNN and HTC, resulting in limited availability of training samples.

\begin{figure}[t]
  \centering
  \includegraphics[width=0.9\columnwidth]{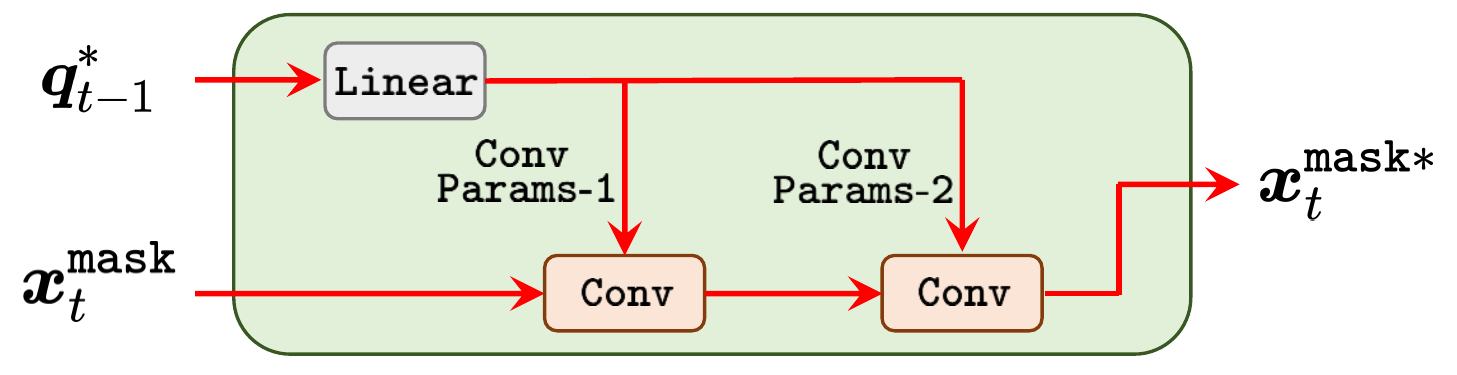}
  \caption{Illustrations of $\mathrm{DynConv}^{\mathtt{mask}}_{t}$ at stage $t$. 
  $\boldsymbol{x}_{t}^{\mathtt{mask*}}$ is enhanced by two consecutive conv-layers, whose kernel parameters are produced by $\boldsymbol{q}_{t - 1}^*$.
  }
  \label{fig: DynConv}
\end{figure}

\subsubsection{Dynamic Mask Head}
Our goal is to design a mask prediction head tailor-made for query based instance segmentation frameworks.
To this end, we propose to leverage dynamic mask heads driven by parallel supervision to replace the vanilla design in Sec.~\ref{sec: vanilla mask head}.
The dynamic mask head at stage $t$ consists of a dynamic mask convolution module $\mathrm{DynConv}^{\mathtt{mask}}_{t}$ (see Fig.~\ref{fig: DynConv})~\cite{SparseRCNN} following by a vanilla mask head $\mathcal{M}_{t}$~\cite{MaskRCNN}.
The mask generation pipeline is reformulated as follows:

\begin{equation}
\begin{aligned}
    \boldsymbol{x}_{t}^{\mathtt{mask}} \leftarrow & \ \mathcal{P}^{\mathtt{mask}}\left(\boldsymbol{x}^{\mathtt{FPN}}, \boldsymbol{b}_{t}\right), \\
    \boldsymbol{x}_{t}^{\mathtt{mask*}} \leftarrow & \ \mathrm{DynConv}^{\mathtt{mask}}_{t} \left(\boldsymbol{x}_{t}^{\mathtt{mask}}, \boldsymbol{q}_{t - 1}^* \right), \\
    \boldsymbol{m}_{t} \leftarrow & \ \mathcal{M}_{t}\left(\boldsymbol{x}_{t}^{\mathtt{mask*}}\right).
\end{aligned}
\end{equation}

It is noteworthy that the only difference between the proposed dynamic mask head and vanilla mask head is the existence of $\mathrm{DynConv}^{\mathtt{mask}}_{t}$. 
We demonstrate that $\mathrm{DynConv}^{\mathtt{mask}}_{t}$ enables $(1)$ per-mask information flow in queries driven by parallel mask branch supervision, and $(2)$ communication and synergy for joint detection and instance segmentation in the following two sub-sections, respectively.
The effectiveness of these two properties is verified in our experiments.

\subsection{Per-mask Information Flow with Parallel Supervision}

In query based models such as~\cite{DETR, DefDETR, SparseRCNN}, the model learns different specialization for each query slot~\cite{DETR}, \ie, $\boldsymbol{q}_{t}[s]$ is the transformed and refined version of previous stage $\boldsymbol{q}_{t - 1}[s]$ in the same $s$-th slot.
Moreover, $\boldsymbol{x}^{\mathtt{mask}}_{t}[s]$ corresponds to and is refined by $\boldsymbol{q}_{t}[s]$~\cite{SparseRCNN}.
Therefore there is an one-to-one correspondence across different stage queries inherent in these frameworks, as well as one-to-one correspondence between mask RoI features and object queries in the same stage.

QueryInst is driven by parallel supervision on dynamic mask heads, which fully leverages the intrinsic one-to-one correspondence in object queries across different stages.
Specifically, we set up dynamic mask heads in parallel with each other, which transform each mask RoI feature $\boldsymbol{x}_{t}^{\mathtt{mask}}$ adaptively in $\mathrm{DynConv}^{\mathtt{mask}}_{t}$ according to the corresponding query $\boldsymbol{q}_{t - 1}^*$, and are simultaneously trained in all stages.
Inside $\mathrm{DynConv}^{\mathtt{mask}}_{t}$, the query acts as memory and is read by mask RoI features $\boldsymbol{x}_{t}^{\mathtt{mask}}$ in the forward pass and written by $\boldsymbol{x}_{t}^{\mathtt{mask}}$ in the backward pass.

During training, the per-mask information (\ie, the mask gradient) not only flows back to mask RoI features $\boldsymbol{x}_{t}^{\mathtt{mask}}$, but also to the object query $\boldsymbol{q}_{t - 1}^*$, which is intrinsically one-to-one interlinked in different stages.
Therefore the per-mask information flow is naturally established by leveraging the inherent properties of query based frameworks, with no additional connection needed.
After the training is completed, the information for mask prediction is stored in queries.

During inference, we throw away all dynamic mask heads in $5$ intermediate stages and only use the final stage predictions for inference.
The queries implicitly carry the multi-stage information for mask prediction, which is read by mask RoI features $\boldsymbol{x}_{t}^{\mathtt{mask}}$ in dynamic mask convolution $\mathrm{DynConv}^{\mathtt{mask}}_{t}$ at the last stage for final mask generation.

Without $\mathrm{DynConv}^{\mathtt{mask}}_{t}$, the link between mask RoI features and the query is lost, and mask heads in different stages are isolated.
Even though parallel supervision is applied to all mask heads, the information related to mask generation cannot flow into queries.
In this condition, QueryInst degenerates to Cascade Mask R-CNN with a fixed number (\ie, $N$) of proposals across all stages.

\subsection{Shared Query and MSA for Joint Detection and Segmentation}
\label{sec: shared query and msa}
At stage $t$, a multi-head self-attention $\mathrm{MSA}_t$ is applied to the query $\boldsymbol{q}_{t - 1}$.
$\mathrm{MSA}_t$ projects the query $\boldsymbol{q}_{t - 1}$ to a high dimensional embedding space and its output $\boldsymbol{q}_{t - 1}^{*}$ is read by the dynamic box convolution $\mathrm{DynConv}^{\mathtt{box}}_{t}$ and dynamic mask convolution $\mathrm{DynConv}^{\mathtt{mask}}_{t}$ respectively, to enhance the task-specific features $\boldsymbol{x}_{t}^{\mathtt{box}}$ and $\boldsymbol{x}_{t}^{\mathtt{mask}}$. 

Throughout this process, the query and $\mathrm{MSA}$ are shared between detection and instance segmentation tasks. 
Both detection and segmentation information flow back into the query through $\mathrm{MSA}$.
This task-joint paradigm establishes a kind of communication and synergy between detection and segmentation tasks, which encourages these two tasks to benefits from each other.
The query learns a better instance-level representation with the guidance of two highly correlated tasks. 
We observe a performance decrease using separate queries or $\mathrm{MSA}$ in our experiments.

\subsection{Comparisons with Cascade Mask R-CNN and HTC}

Cascade Mask R-CNN~\cite{CascadeRCNN_TPAMI} presents a multi-stage architecture to resample with higher intersection over union (IoU) thresholds for the latter stages, and progressively refine the training distributions of region proposals~\cite{CascadeRCNN, CascadeRCNN_TPAMI}.
This resampling operation guarantees the availability of a large number of accurate localized proposals for the final stage.
Despite outperforming non-cascade counterparts, the effectiveness of Cascade Mask R-CNN mainly stems from the progressively refined proposal recall.
Whereas, the mask heads across different stages are isolated, and the input feature for mask head always come from the same FPN feature regardless of the stage.

To mitigate the aforementioned drawback of Cascade Mask R-CNN, HTC~\cite{HTC} improves Cascade Mask R-CNN by introducing direct and explicit connections across the mask heads at different stages.
The current stage mask features are combined with the accumulated mask features from all previous stages.
Equivalently, the mask head of the final stage is $3 \times$ deeper than the first stage.
Therefore the final stage mask prediction can benefit from deeper features. 
Establishing direct mask information flow as HTC can alleviate the issue in Cascade Mask R-CNN to some extent.
However, this explicit connection across mask heads at different R-CNN stages results in inefficient training and inference.

There are some potential issues inherent in the aforementioned non-query based instance segmentation paradigms.
For Cascade Mask R-CNN and HTC, the quality of proposals in different stages is refined \textit{in the statistical sense}~\cite{CascadeRCNN, CascadeRCNN_TPAMI, HTC}.
For each stage, the number and distribution of training samples are quite different, and there is no explicit and intrinsic correspondence for each individual proposal across different stages~\cite{CenterNet2}.
Moreover, there is also a mismatch between the training and inference sample distribution~\cite{SCNet}.
Therefore, introducing direct connections at the architecture-level is necessary for the mask heads in different stages to explicitly learn the correspondence~\cite{HTC}.

Our method does not directly solve the aforementioned issues, but bypasses them.
For QueryInst, the connections across stages are naturally established by one-to-one correspondence inherent in queries.
This approach eliminates the explicit multi-stage mask head connection and the proposal distribution inconsistency issues.
We show that the proposed new paradigm can surpass Cascade Mask R-CNN and HTC in terms of both accuracy and speed.


\subsection{QueryInst-VIS for Video Instance Segmentation}
\label{section: online video instance segmentation}

Video instance segmentation (VIS)~\cite{VIS} is a highly relevant task to still-image instance segmentation that aims at detecting, classifying, segmenting and tracking visual instances over video frames. 
We demonstrate that QueryInst can be easily extended to VIS with minimal modifications by simply adding the vanilla track head in MaskTrack R-CNN baseline~\cite{VIS}.
The proposed model coined as QueryInst-VIS can perform video instance segmentation in an online manner while operating at real-time.
The total training and inference pipeline keep the same as MaskTrack R-CNN. 
We evaluate QueryInst-VIS on the challenging YouTube-VIS~\cite{VIS} benchmark to demonstrate its effectiveness.

\begin{table*}[t]
  \centering
  \setlength{\tabcolsep}{1.5 pt}
  \begin{tabular}{l|c|c|c|c|c|cc|ccc|r}
    \hline
    
    \hline

    \hline
    
     \rowcolor{mygray}
       Method
     & Backbone
     & Aug.
     & Epochs
     & AP$^{\mathtt{box}}$
     & AP
     & AP$_{50}$
     & AP$_{75}$
     & AP$_{S}$
     & AP$_{M}$
     & AP$_{L}$
     & FPS
     \\
    \hline
    \hline
    Mask R-CNN~\cite{MaskRCNN}
    & \multirow{4}{*}{ResNet-$50$-FPN} 
    & \multirow{4}{*}{$640 \sim 800$} 
    & \multirow{4}{*}{$36$} 
    & $41.3$ 
    & $37.5$ 
    & $59.3$ & $40.2$ & $21.1$ & $39.6$ & $48.3$ 
    & $14.0$
    \\
    CondInst w/ sem.~\cite{CondInst_TPAMI}
    &  &  &  
    &  --
    & $38.6$ 
    & $60.2$ & $41.4$ & $20.6$ & $41.0$ & $51.1$ 
    & $14.1$
    \\
    SOLOv2~\cite{SOLO}
    &  &  &  
    & $40.4$
    & $38.8$ 
    & $59.9$ & $41.7$ & $16.5$ & $\mathbf{41.7}$ & $\mathbf{56.2}$ 
    & $13.8$
    \\
    \textbf{QueryInst} ($5$ Stage, $100$ Queries)
    &  &  & 
    & $\mathbf{44.5}$ 
    & $\mathbf{39.9}$ 
    & $\mathbf{62.2}$ & $\mathbf{43.0}$ & $\mathbf{22.9}$ & $\mathbf{41.7}$ & $51.9$ 
    & $13.5$
    \\
    
    \hline
    
    Cascade Mask R-CNN~\cite{CascadeRCNN_TPAMI}
    & \multirow{4}{*}{ResNet-$50$-FPN} 
    & \multirow{4}{*}{$640 \sim 800$} 
    & \multirow{4}{*}{$36$} 
    & $44.5$ 
    & $38.6$ 
    & $60.0$ & $41.7$ & $21.7$ & $40.8$ & $49.6$ 
    & $10.4$
    \\
    HTC~\cite{HTC}
    &  &  & 
    & $44.9$ 
    & $39.7$ 
    & $61.4$ & $43.1$ & $22.6$ & $42.2$ & $50.6$ 
    & $3.1$
    \\
    \textbf{QueryInst} ($100$ Queries)
    &  &  & 
    & $44.8$ 
    & $40.1$ 
    & $62.3$ & $43.4$ & $23.3$ & $42.1$ & $52.0$ 
    & $10.5$
    \\
    \textbf{QueryInst} ($300$ Queries)
    &  &  & 
    & $\mathbf{45.6}$ 
    & $\mathbf{40.6}$ 
    & $\mathbf{63.0}$ & $\mathbf{44.0}$ & $\mathbf{23.4}$ & $\mathbf{42.5}$ & $\mathbf{52.8}$ 
    & $7.0$
    \\
    \hline
    Cascade Mask R-CNN
    & \multirow{3}{*}{ResNet-$101$-FPN} 
    & \multirow{3}{*}{$640 \sim 800$} 
    & \multirow{3}{*}{$36$} 
    & $45.7$ 
    & $39.8$ 
    & $61.6$ & $43.0$ & $22.4$ & $42.2$ & $50.8$ 
    & $8.7$
    \\
    HTC
    &  &  & 
    & $46.2$ 
    & $40.7$ 
    & $62.7$ & $44.2$ & $23.1$ & $43.4$ & $52.7$ 
    & $2.5$
    \\
    \textbf{QueryInst} ($300$ Queries)
    &  &  & 
    & $\mathbf{47.0}$ 
    & $\mathbf{41.7}$ 
    & $\mathbf{64.4}$ & $\mathbf{45.3}$ & $\mathbf{24.2}$ & $\mathbf{43.9}$ & $\mathbf{53.9}$ 
    & $6.1$
    \\
    \hline
    Cascade Mask R-CNN
    & \multirow{4}{*}{ResNet-$101$-FPN} 
    & \multirow{4}{*}{\shortstack[c]{$480 \sim 800$ \\ w/ crop}} 
    & \multirow{4}{*}{$36$} 
    & $46.2$ 
    & $40.0$ 
    & $61.7$ & $43.5$ & $22.5$ & $42.5$ & $51.2$ 
    & $8.7$
    \\
    HTC
    &  &  & 
    & $46.3$ 
    & $40.8$ 
    & $62.6$ & $44.3$ & $23.0$ & $43.5$ & $52.6$ 
    & $2.5$
    \\
    Sparse R-CNN ($300$ Queries)
    &  &  & 
    & $46.3$ 
    & $-$ 
    & $-$ & $-$ & $-$ & $-$ & $-$
    & $6.9$
    \\
    \textbf{QueryInst} ($300$ Queries)
    &  &  & 
    & $\mathbf{48.1}$ 
    & $\mathbf{42.8}$ 
    & $\mathbf{65.6}$ & $\mathbf{46.7}$ & $\mathbf{24.6}$ & $\mathbf{45.0}$ & $\mathbf{55.5}$ 
    & $6.1$
    \\
    \hline
    \multirow{2}{*}{\shortstack[c]{\textbf{QueryInst} ($300$ Queries)}}
    & \multirow{2}{*}{\shortstack[c]{ResNeXt-$101$-FPN \\ w/ DCN}}
    & \multirow{2}{*}{\shortstack[c]{$480 \sim 800$ \\ w/ crop}} 
    & \multirow{2}{*}{$36$} 
    & \multirow{2}{*}{\shortstack[c]{$\mathbf{50.4}$}}
    & \multirow{2}{*}{\shortstack[c]{$\mathbf{44.6}$}} 
    & \multirow{2}{*}{\shortstack[c]{$\mathbf{68.1}$}} 
    & \multirow{2}{*}{\shortstack[c]{$\mathbf{48.7}$}} 
    & \multirow{2}{*}{\shortstack[c]{$\mathbf{26.6}$}} 
    & \multirow{2}{*}{\shortstack[c]{$\mathbf{46.9}$}} 
    & \multirow{2}{*}{\shortstack[c]{$\mathbf{57.7}$}}
    & \multirow{2}{*}{$3.1$}
    \\
    & & & & & & & & & & &
    \\
    \hline
    \multirow{2}{*}{\shortstack[c]{\textbf{QueryInst} ($300$ Queries) @ $\mathtt{val}$}}
    & \multirow{2}{*}{\shortstack[c]{Swin-L}}
    & \multirow{2}{*}{\shortstack[c]{$400 \sim 1200$ \\ w/ crop}} 
    & \multirow{2}{*}{$50$} 
    & \multirow{2}{*}{\shortstack[c]{$\mathbf{56.1}$}}
    & \multirow{2}{*}{\shortstack[c]{$\mathbf{48.9}$}} 
    & \multirow{2}{*}{\shortstack[c]{$\mathbf{74.0}$}} 
    & \multirow{2}{*}{\shortstack[c]{$\mathbf{53.9}$}} 
    & \multirow{2}{*}{\shortstack[c]{$\mathbf{30.8}$}} 
    & \multirow{2}{*}{\shortstack[c]{$\mathbf{52.6}$}} 
    & \multirow{2}{*}{\shortstack[c]{$\mathbf{68.3}$}}
    & \multirow{2}{*}{$3.3^{\top}$}
    \\
    & & & & & & & & & & &
    \\
    \hline
    \multirow{2}{*}{\shortstack[c]{\textbf{QueryInst} ($300$ Queries)}}
    & \multirow{2}{*}{\shortstack[c]{Swin-L}}
    & \multirow{2}{*}{\shortstack[c]{$400 \sim 1200$ \\ w/ crop}} 
    & \multirow{2}{*}{$50$} 
    & \multirow{2}{*}{\shortstack[c]{$\mathbf{56.1}$}}
    & \multirow{2}{*}{\shortstack[c]{$\mathbf{49.1}$}} 
    & \multirow{2}{*}{\shortstack[c]{$\mathbf{74.2}$}} 
    & \multirow{2}{*}{\shortstack[c]{$\mathbf{53.8}$}} 
    & \multirow{2}{*}{\shortstack[c]{$\mathbf{31.5}$}} 
    & \multirow{2}{*}{\shortstack[c]{$\mathbf{51.8}$}} 
    & \multirow{2}{*}{\shortstack[c]{$\mathbf{63.2}$}}
    & \multirow{2}{*}{$3.3^\top$}
    \\
    & & & & & & & & & & &
    \\
    \hline
    
    \hline
    
    \hline
  \end{tabular}
  \smallskip
  \caption{Main results on COCO $\mathtt{test}$-$\mathtt{dev}$. 
  The numbers under ``Aug." indicate the scale range of the shorter size of inputs with a stride of $32$. 
  AP$^{\mathtt{box}}$ denotes box AP.
  AP without superscript denotes mask AP.
  The best results are in \textbf{bold} for each configuration. Superscript ``$\top$" indicates the FPS data are measured on a single RTX $2080$Ti GPU with batch size $1$.}
  \label{tab: main results on test-dev}
\end{table*}

\section{Experiments}

\subsection{Dataset}

\noindent
\textbf{COCO.}
Most of our experiments are conducted on the challenging COCO dataset~\cite{COCO}. 
Following the common practice, we use the COCO $\mathtt{train2017}$ split ($115k$ images) for training and the $\mathtt{val2017}$ split ($5k$ images) as validation for our ablation study. 
We report our main results on the $\mathtt{test}$-$\mathtt{dev}$ split ($20k$ images).

\noindent
\textbf{Cityscapes.}
Cityscapes~\cite{Cityscapes} is an ego-centric street-scene dataset with $8$ categories, $2975$ train images, and $500$ validation images for instance segmentation. 
The images are with higher resolution ($1024 \times 2048$ pixels) compared with COCO, and have more pixel-accurate ground-truth.

\noindent
\textbf{YouTube-VIS.}
In addition to static-image instance segmentation, we demonstrate the effectiveness of our QueryInst on video instance segmentation.
YouTube-VIS~\cite{VIS} is a challenging dataset for video instance segmentation task, which has a $40$-category label set, $4,883$ unique video instances and $131k$ high-quality manual annotations.
There are $2,238$ training videos, $302$ validation videos, and $343$ test videos.

\subsection{Implementation Details}

\noindent
\textbf{Training Setup.}
Our implementation is based on {MMDetection}~\cite{mmdetection} and {Detectron2}~\cite{Detectron2}.
Following~\cite{SparseRCNN}, 
the default training schedule is $36$ epochs and the initial learning rate is set to $2.5 \times 10^{-5}$, divided by $10$ at $27$-th epoch and $33$-th epoch, respectively.
We adopt AdamW optimizer with $1 \times 10^{-4}$ weight decay.
Hyper-parameters, configurations as well as the label assignment procedures follow the setting in~\cite{DETR, DefDETR, SparseRCNN}.
In total, the R-CNN head of QueryInst contains $6$ stages in parallel as~\cite{SparseRCNN}. 
The mask head is trained by minimizing dice loss~\cite{DiceLoss}.
Without special mentioning, we adopt QueryInst model trained with $100$ queries and ResNet-$50$-FPN~\cite{ResNet, FPN} as backbone in our experiments in the ablation studies. 

\noindent
\textbf{Inference.}
Given an input image, QueryInst directly outputs top $100$ bounding box predictions with their scores and corresponding instance masks without further post-processing.
For inference, we use the final stage masks as the predictions and ignore all the parallel $\mathrm{DynConv}^{\mathtt{mask}}$ at the intermediate stages.
The inference speed reported is measured using a single Titan Xp GPU with inputs resized to have their shorter side being $800$ and their longer side less or equal to $1333$.

\begin{table*}[t]
  \centering
  \setlength{\tabcolsep}{3.0 pt}
  \begin{tabular}{l|c|c|cc|cccccccc}
    \hline
    
    \hline

    \hline
    
    \rowcolor{mygray}
  Method
  & Backbone
  & AP$_{\mathtt{val}}$
  & AP
  & AP$_{50}$
  & person
  & rider
  & car
  & trunk
  & bus
  & train
  & mcycle
  & bicycle
  \\
  \hline
  \hline
  Mask R-CNN~\cite{MaskRCNN} & ResNet-$50$  & $36.4$ & $32.0$ & $58.1$ & $34.8$ & $27.0$ & $49.1$ & $30.1$ & $40.9$ & $30.9$ & $24.1$ & $18.7$ \\
  BShapeNet+~\cite{BShapeNet+} & ResNet-$50$  & $-$ & $32.9$ & $58.8$ & $36.6$ & $24.8$ & $50.4$ & $33.7$ & $41.0$ & $33.7$ & $25.4$ & $17.8$ \\
  UPSNet~\cite{UPSNet} & ResNet-$50$  & $37.8$ & $33.0$ & $\mathbf{59.7}$ & $35.9$ & $27.4$ & $51.9$ & $31.8$ & $43.1$ & $31.4$ & $23.8$ & $19.1$ \\
  CondInst~\cite{CondInst_TPAMI} & ResNet-$50$  & $37.5$ & $33.2$ & $57.2$ & $35.1$ & $27.7$ & $54.5$ & $29.5$ & $42.3$ & $\mathbf{33.8}$ & $23.9$ & $18.9$ \\
  CondInst~\cite{CondInst_TPAMI} w/ sem. & DCN-$101$-BiFPN  & $39.3$ & $33.9$ & $58.2$ & $35.6$ & $28.1$ & $55.0$ & $\mathbf{32.1}$ & $\mathbf{44.2}$ & $33.6$ & $24.5$ & $18.6$ \\
  \hline
  \textbf{QueryInst} & ResNet-$50$  & $\mathbf{39.4}$ & $\mathbf{34.4}$ & $59.6$ & $\mathbf{40.4}$ & $\mathbf{30.7}$ & $\mathbf{56.8}$ & $29.1$ & $40.5$ & $30.8$ & $\mathbf{26.0}$ & $\mathbf{21.1}$ \\
  \hline

  \hline
  
  \hline
  \end{tabular}
  \smallskip
  \caption{Instance segmentation results on Cityscapes $\mathtt{val}$ (AP$_{\mathtt{val}}$ column) and $\mathtt{test}$ (remain columns) split.
  The best results are in \textbf{bold}.}
  \label{tab: cityscapes results}
\end{table*}

\begin{table*}[t]
  \centering
  \setlength{\tabcolsep}{13.5 pt}
  \begin{tabular}{l|c|c|cc|cc|c}
    \hline
    
    \hline

    \hline
    
    \rowcolor{mygray}
  Method
  & Backbone  
  & AP
  & AP$_{50}$
  & AP$_{75}$
  & AR$_{1}$
  & AR$_{10}$
  & FPS
  \\
  \hline
  \hline
  MaskTrack R-CNN~\cite{VIS} & ResNet-$50$ 
  & $30.3$ & $51.1$ & $32.6$ & $31.0$ & $35.5$ & $22.1$ \\
  SipMask-VIS~\cite{SipMask} & ResNet-$50$ 
  & $32.5$ & $53.0$ & $33.3$ & $33.5$ & $38.9$ & $30.9$ \\
  SipMask-VIS$^*$ & ResNet-$50$ 
  & $33.7$ & $54.1$ & $35.8$ & $35.4$ & $40.1$ & $30.9$ \\
  STEm-Seg~\cite{STEm-Seg} & ResNet-$50$ 
  & $30.6$ & $50.7$ & $33.5$ & $31.6$ & $37.1$ & $4.4$  \\
  STEm-Seg & ResNet-$101$ 
  & $34.6$ & $55.8$ & $37.9$ & $34.4$ & $41.6$ & $2.1$  \\
  CompFeat~\cite{CompFeat} & ResNet-$50$ 
  & $35.3$ & $56.0$ & $38.6$ & $33.1$ & $40.3$ & --     \\
  VisTR~\cite{VisTR} & ResNet-$50$ 
  & $34.4$ & $55.7$ & $36.5$ & $33.5$ & $38.9$ & $30.0$ \\
  VisTR & ResNet-$101$ 
  & $35.3$ & $\mathbf{57.0}$ & $36.2$ & $34.3$ & $40.4$ & $27.7$ \\
  \hline
  \textbf{QueryInst-VIS} & ResNet-$50$ 
  & $34.6$ & $55.8$ & $36.5$ & $35.4$ & $42.4$ & $\mathbf{32.3}$ \\
  \textbf{QueryInst-VIS$^*$} & ResNet-$50$ 
  & $\mathbf{36.2}$ & $56.7$ & $\mathbf{39.7}$ & $\mathbf{36.1}$ & $\mathbf{42.9}$ & $\mathbf{32.3}$ \\
  \hline

  \hline
  
  \hline
  \end{tabular}
  \smallskip
  \caption{Comparisons with state-of-the-art video instance segmentation methods on YouTube-VIS $\mathtt{val}$ set. 
  Methods with superscript ``$*$" indicates using multi-scale data argumentation during training.
  The best results are in \textbf{bold}.}
  \label{tab: YouTubeVIS Results}
\end{table*}

\subsection{Main Results}

\noindent
\textbf{Comparisons on COCO Instance Segmentation.}

\noindent
The comparison of QueryInst with the state-of-the-art instance segmentation methods on COCO $\mathtt{test}$-$\mathtt{dev}$ are listed in Tab.~\ref{tab: main results on test-dev}.
We have tested different backbones and data augmentations.
CondInst~\cite{CondInst_TPAMI} (with auxiliary semantic branch) and SOLOv2~\cite{SOLOv2} are the latest state-of-the-art instance segmentation approach based on dynamic convolutions.
A $5$-stage QueryInst trained with $100$ queries outperforms them with over $1.1$ mask AP gain under similar inference speed.
QueryInst trained with $100$ queries can also surpass Cascade Mask R-CNN~\cite{MaskRCNN} by $1.5$ mask AP while runs with the same FPS.
For fair comparisons with HTC~\cite{HTC}, we train HTC using the $36$ epochs training schedule and multi-scale data augmentations following the standard setting in~\cite{MaskRCNN, Detectron2}, yielding $\sim 1$ higher mask AP than original results reported in~\cite{HTC}.
Under same experimental conditions, QueryInst outperforms the state-of-the-art HTC in terms of both accuracy and speed.
Moreover, QueryInst outperforms HTC in terms of AP at different IoU thresholds (AP$_{50}$ and AP$_{75}$) as well as AP at different scales (AP$_{S}$, AP$_{M}$ and AP$_{L}$), regardless of the experimental configuration.
We also find that compared with Cascade Mask R-CNN and HTC, the query based QueryInst can benefit more from stronger data argumentation used in~\cite{DETR, DefDETR, SparseRCNN}\footnote{We experimentally study the effects of different training schedules and data augmentations to the Mask R-CNN family in the Appendix.}.
Specifically, using ResNet-$101$-FPN~\cite{FPN} backbone and stronger multi-scale data argumentation with random crop, QueryInst surpasses HTC by $2.0$ mask AP and $1.8$ box AP while runs $2.4 \times$ faster.
Further, QueryInst with deformable ResNeXt-$101$-FPN backbone~\cite{ResNeXt, DCNv1, DCNv2} achieves $44.6$ mask AP and $50.4$ box AP without bells and whistles.

We demonstrate that the instance segmentation performance of QueryInst is not simply come from the accurate bounding box provided by Sparse R-CNN~\cite{SparseRCNN} object detector.
On the contrary, QueryInst can largely improve the detection performance.
The best result of Sparse R-CNN (ResNet-$101$-FPN, $300$ queries, $480 \sim 800$ w/ crop, $36$ epochs) reported in~\cite{mmdetection} is $46.3$ box AP.
Under the same experimental setting, QueryInst can achieve $48.1$ box AP, which outperform Sparse R-CNN by $1.8$ box AP.
We also show in the ablation study that QueryInst can outperform Cascade Mask R-CNN and HTC based on a weaker query based detector.

We also apply QueryInst to the recent state-of-the-art Swin Transformer~\cite{swin} backbone without further modifications, and we find the proposed model is quite capable of adapting with Swin-L.
Without bells and whistles, QueryInst can achieve the art performance in instance segmentation\footnote{\url{https://paperswithcode.com/sota/instance-segmentation-on-coco}} as well as object detection\footnote{\url{https://paperswithcode.com/sota/object-detection-on-coco}}.
For the first time, we demonstrate that an end-to-end query based framework driven by parallel supervision is competitive with well-established and highly-optimized methods in instance-level recognition tasks.

\noindent
\textbf{Comparisons on Cityscapes Instance Segmentation.}

\noindent
We also conduct experiments on Cityscapes dataset to demonstrate the generalization of QueryInst.
Following the standard setting in~\cite{CondInst_TPAMI, MaskRCNN}, all models are first pre-trained on COCO $\mathtt{train2017}$ split then finetuned on Cityscapes using $\mathtt{fine}$ annotations for $24k$ iterations with batch size $8$  ($1$ image per GPU).
The initial learning rate is linearly scaled to $1.25 \times 10^{-5}$ and is reduced by a factor of $10$ at step $18k$.

The results are shown in Tab.~\ref{tab: cityscapes results}.
QueryInst achieves $39.4$ AP on $\mathtt{val}$ split and $34.4$ AP on $\mathtt{test}$ split, surpassing several strong baselines.
Notably, compared to the dynamic convolution based method CondInst~\cite{CondInst_TPAMI}, QueryInst with ResNet-$50$ backbone outperforms CondInst with both ResNet-$101$-DCN-BiFPN backbone and semantic branch. 
Overall, our QueryInst achieves leading results on Cityscapes dataset without bells and whistles.

\begin{table*}[t]
  \centering
  \setlength{\tabcolsep}{3.0 pt}
  \begin{tabular}{c|ccc|c|cc|cc|rc}
    \hline
    
    \hline

    \hline
    
    \rowcolor{mygray}
     & & & & & & & & & &
     \\
     
    \rowcolor{mygray}
       \multirow{-2}{*}{\shortstack[c]{Type}}
     & \multirow{-2}{*}{\shortstack[c]{Cascade \\ Mask Head~\cite{CascadeRCNN_TPAMI}}}
     & \multirow{-2}{*}{\shortstack[c]{HTC \\ Mask Flow~\cite{HTC}}}
     & \multirow{-2}{*}{\shortstack[c]{\color{blue} \color{blue}$\mathrm{DynConv}^{\mathtt{mask}}$}}
     & \multirow{-2}{*}{\shortstack[c]{Fig.}}
     & \multirow{-2}{*}{\shortstack[c]{AP$^{\mathtt{box}}$}}
     & \multirow{-2}{*}{\shortstack[c]{$\Delta^{\mathtt{box}}$}}
     & \multirow{-2}{*}{\shortstack[c]{AP$^{\mathtt{mask}}$}}
     & \multirow{-2}{*}{\shortstack[c]{$\Delta^{\mathtt{mask}}$}}
     & \multirow{-2}{*}{\shortstack[c]{FPS}}
     & \multirow{-2}{*}{\shortstack[c]{$\Delta^{\mathtt{FPS}}$}}
     \\

     \hline\hline
     \multirow{2}{*}{\shortstack[c]{Non-query Based}} 
     & \checkmark & & & & $44.3$ & & $38.5$ & & $10.4$ & \\
     & & \checkmark & & & $44.4$ & $ + \ 0.1$ & $39.3$ & $ + \ 0.8$ & $3.1$ & $ - \ 7.3$\\
     \hline
     \multirow{4}{*}{\shortstack[c]{Query Based}}
     & \checkmark & & & Fig.~\ref{fig: QueryInst} (b) & $43.8$ & & $37.9$ & & $11.1$ & \\
     & & \checkmark & & & $43.8$ & $ + \ 0.0$ & $38.9$ & $ + \ 1.0$ & $6.0$ & $ - \ 5.1$ \\
     & & & {\color{blue}\checkmark} & {\color{blue}Fig.~\ref{fig: QueryInst} (c)} & {\color{blue}$44.5$} & {\color{blue}$ + \ 0.7$} & {\color{blue}$39.8$} & {\color{blue}$ + \ 1.9$} & {\color{blue}$10.5$} & {\color{blue}$ - \ 0.6$} \\
     & & \checkmark & \checkmark & & $44.4$ & $ + \ 0.6$ & $40.0$ & $ + \ 2.1$ & $5.4$ & $ - \ 5.7$ \\
     \hline
    
     \hline
     
     \hline
  \end{tabular}
  \smallskip
  \caption{Impacts of different mask head architectures on different frameworks. The setting in {\color{blue} blue} is our default instantiation.}
  \label{tab: mask head ablation}
\end{table*}

\noindent
\textbf{Video Instance Segmentation Results on YouTube-VIS.}

\noindent
Tab.~\ref{tab: YouTubeVIS Results} shows the video instance segmentation results on YouTube-VIS $\mathtt{val}$ set.
Following the standard setting in~\cite{VIS}, we first pre-train the instance segmentation model on COCO $\mathtt{train2017}$, then we finetune the corresponding VIS model on YouTube-VIS $\mathtt{train}$ set for $12$ epochs. 
The maximum number of instances in one frame in YouTube-VIS dataset is $10$, so we set the number of queries to $10$ in QueryInst-VIS.
The setting enables the model to operate at real-time ($> 30$ FPS). 

As mentioned in Sec.~\ref{section: online video instance segmentation}, QueryInst-VIS adopts the vanilla track method of MaskTrack R-CNN~\cite{VIS} and SipMask-VIS~\cite{SipMask}, while it obtains $4.3$ AP improvement compared to MaskTrack R-CNN and $2.1$ AP improvement compared to SipMask-VIS. 
Moreover, QueryInst can outperform many well-established and highly-optimized VIS approaches, such as STEm-Seg, CompFeat and VisTR in terms of both accuracy and speed.

    
     
     
     
     
     
    

\begin{table}[t]
  \centering
  \setlength{\tabcolsep}{3.40 pt}
  \begin{tabular}{cc|c|c|c|c}
    \hline
    
    \hline

    \hline
    
    \rowcolor{mygray}
       Parallel
     & $\mathrm{DynConv}^{\mathtt{mask}}$ 
     & Fig.
     & AP$^{\mathtt{box}}$
     & AP$^{\mathtt{mask}}$
     & FPS
     \\
     \hline\hline
     & & {Fig.~\ref{fig: QueryInst} (b)} & $43.5$ & $37.4$ & $11.1$ \\
     
     \checkmark & & {Fig.~\ref{fig: QueryInst} (b)} & $43.8$ & $37.9$ & $11.1$  \\

     & \checkmark & {Fig.~\ref{fig: QueryInst} (c)} & $43.8$ & $38.8$ & $10.5$ \\
     
     \checkmark & \checkmark & {Fig.~\ref{fig: QueryInst} (c)} & {$44.5$} & {$39.8$} & {$10.5$} \\
     
     \hline
    
     \hline
     
     \hline
  \end{tabular}
  \smallskip
  \caption{Impacts of parallel supervision and $\mathrm{DynConv}^{\mathtt{mask}}$.}
  \label{tab: component-wise analysis}
\end{table}

\begin{table}[t]
  \centering
  \setlength{\tabcolsep}{2.2 pt}
  \begin{tabular}{cc|c|c|c|c}
    \hline
    
    \hline

    \hline
    
    \rowcolor{mygray}
       Shared MSA
     & Shared Query
     & AP$^{\mathtt{box}}$
     & $\Delta^{\mathtt{box}}$
     & AP$^{\mathtt{mask}}$
     & $\Delta^{\mathtt{mask}}$
     \\
     \hline\hline
     & & $43.4$ & & $38.1$ &\\
     \checkmark & & $43.9$ & $ + \ 0.5$ & $38.3$ & $ + \ 0.2$\\
     & \checkmark & $44.1$ & $ + \ 0.7$ & $39.5$ & $ + \ 1.4$\\
     \checkmark & \checkmark & $44.5$ & $ + \ 1.1$ & $39.8$ & $ + \ 1.7$ \\
     \hline
    
     \hline
     
     \hline
  \end{tabular}
  \smallskip
  \caption{Impacts of using shared query and MSA.}
  \label{tab: queries and MSA ablation}
\end{table}

\subsection{Ablation Study}

\noindent
\textbf{Study of Parallel Supervision and $\mathrm{DynConv}^{\mathtt{mask}}$.}

\noindent
We show that applying parallel mask head supervision and $\mathrm{DynConv}^{\mathtt{mask}}$ are both indispensable for good performance.
As shown in Tab.~\ref{tab: component-wise analysis}, using parallel supervision on vanilla mask head cannot bring large improvement, because mask heads in different stages are isolated and there is no cross-stage per-mask information flow established (Sec.~\ref{sec: vanilla mask head}).
Using $\mathrm{DynConv}^{\mathtt{mask}}$ without parallel supervision on each stage can only bring moderate improvement, for mask gradients injected from the final stage cannot fully driven the per-mask information flow across queries in all stages.
When $\mathrm{DynConv}^{\mathtt{mask}}$ in all stages are driven by parallel supervision simultaneously, QueryInst achieves significant improvement in accuracy with only little drops in inference speed.
The reason is that during inference, we throw away all the parallel $\mathrm{DynConv}^{\mathtt{mask}}$ in the intermediate stage and only use the final stage mask predictions.
The per-mask information is written and preserved in queries during training, which only need to be read out at the final stage during inference.

\noindent
\textbf{Study of Query and MSA.}

\noindent
Tab.~\ref{tab: queries and MSA ablation} studies the impact of using shared query and MSA.
As expected in Sec.~\ref{sec: shared query and msa}, using shared query and MSA simultaneously establishes a kind of communication and synergy between detection and segmentation tasks, which encourages this two tasks to benefits from each other and achieves the highest box AP and mask AP.
Moreover, this configuration consumes minimal parameters and computation budgets.
Therefore we choose using shared query and MSA as the default instantiation of our QueryInst.

\begin{figure}[t]
  \centering
  \includegraphics[width=1.0\columnwidth]{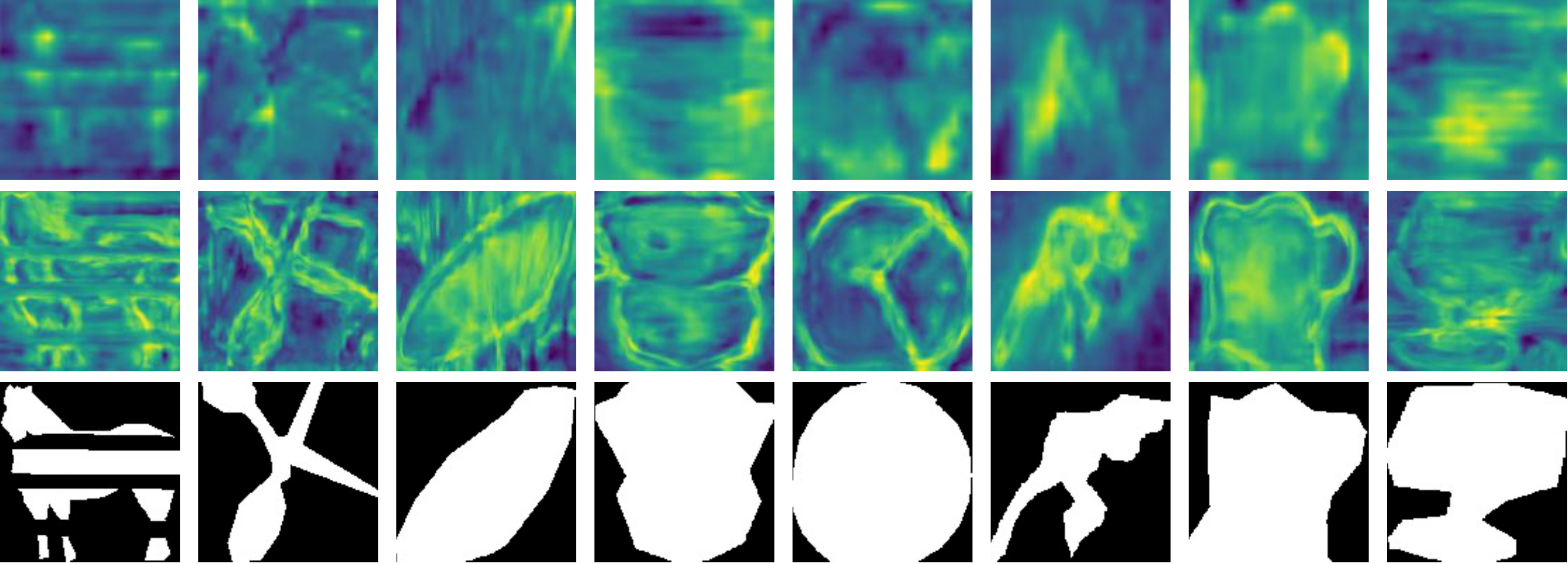}
  \caption{Effects of $\mathrm{DynConv}^{\mathtt{mask}}$.
  The first row shows mask features $\boldsymbol{x}^{\mathtt{mask}}$ directly extracted from FPN.
  The second row shows mask features $\boldsymbol{x}^{\mathtt{mask*}}$ enhanced by queries in $\mathrm{DynConv}^{\mathtt{mask}}$. 
  Last row is ground-truth instance masks. 
  The results show that mask features enhanced by queries yield more genuine and accurate details and carry more information of instances.
  }
  \label{fig: query driven}
\end{figure}

\begin{figure*}[t]
    \centering
    \includegraphics[width=1.9\columnwidth]{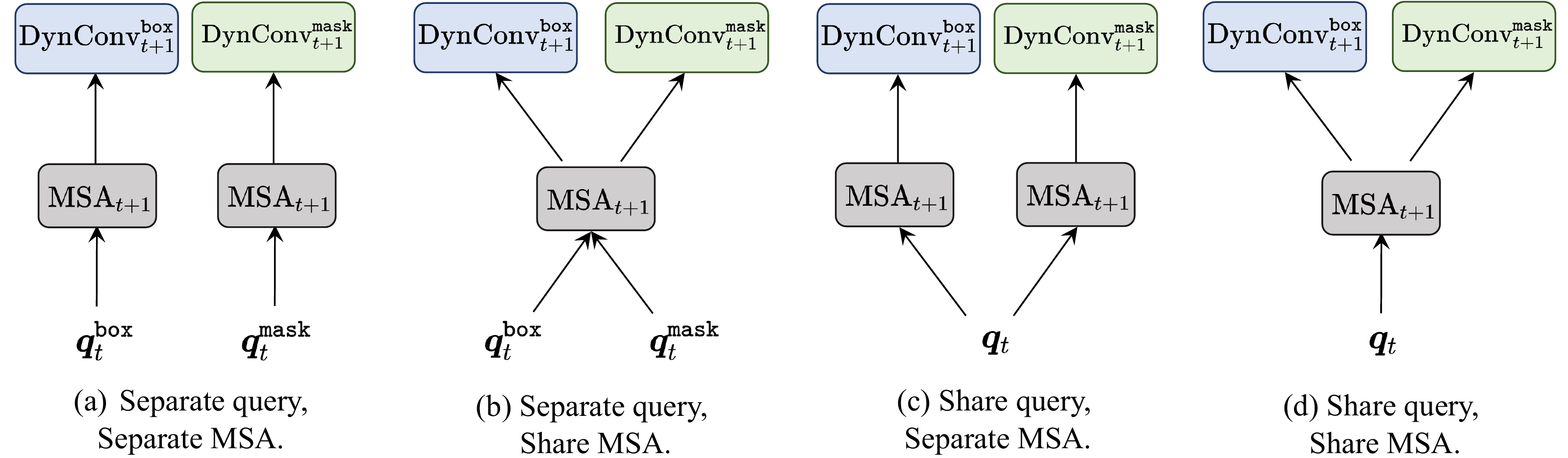}
    \caption{Illustration of $4$ different query and MSA configurations. 
    We use (d) as the default instantiation of QueryInst.}
    \label{fig: share query and share msa.}
\end{figure*}

\begin{table}[t!]
    \centering
    \setlength{\tabcolsep}{4.5 pt}
    \begin{tabular}{l|c|c|c}
    \hline
    
    \hline

    \hline
    
    \rowcolor{mygray}
         Methods & Sched. & Training Time & AP  \\
         \hline
         \hline
         HTC~\cite{HTC} & $3\times$ & $\sim41$ hours & $39.7$ \\
         QueryInst ($100$ Queries) & $3\times$ & $\sim {35}$ hours & $40.1$ \\
         QueryInst ($300$ Queries) & $3\times$ & $\sim {38}$ hours & $40.6$ \\
         \hline
         
         \hline
         
         \hline
    \end{tabular}
    \smallskip
    \caption{Training time comparisons between QueryInst and HTC. 
    All models use ResNet-$50$-FPN~\cite{ResNet, FPN} as backbone and are trained with $3\times$ schedule ($\sim 36$ epochs) on $8$ NVIDIA V$100$ GPUs ($2$ images per GPU).
    QueryInst achieves better performance while needs less training time.}
    \label{tab: training time}
\end{table}

\noindent
\textbf{Study of Different Mask Head.}

\noindent
Tab.~\ref{tab: mask head ablation} studies the impact of different mask head architectures on query and non-query based frameworks.
All stages are simultaneously trained.
For non-query based frameworks, the $1$-st row is the results of Cascade Mask R-CNN~\cite{CascadeRCNN_TPAMI} and the $2$-nd row is HTC~\cite{HTC}.
We have the following $3$ major observations. 

First, we find that directly integrating cascade mask head~\cite{CascadeRCNN_TPAMI} and HTC mask flow~\cite{HTC} into the query based model is not as effective as in its original framework.
When cascade mask head is applied ($3$-th row), the query based model is $0.5$ AP$^\mathtt{box}$ and $0.6$ AP$^\mathtt{mask}$ lower than the original Cascade Mask R-CNN ($1$-th row).
When HTC mask flow is applied ($4$-th row), the query based model is $0.6$ AP$^\mathtt{box}$ and $0.4$ AP$^\mathtt{mask}$ lower than the original HTC ($2$-th row).
These results demonstrate that previous successful empirical practice from non-query based multi-stage models is possibly inadequate for query based models (Sec.~\ref{sec: vanilla mask head}).

Second, when the proposed parallel $\mathrm{DynConv}^{\mathtt{mask}}$ is applied to the query based model, QueryInst ($5$-th row) outperforms the baseline ($3$-th row) by $0.7$ AP$^\mathtt{box}$ and $1.9$ AP$^\mathtt{mask}$, while maintaining a high FPS.
Moreover, QueryInst also beat original HTC ($2$-th row) in terms of both AP$^\mathtt{box}$ and AP$^\mathtt{mask}$ while runs about $3 \times$ faster.
Fig.~\ref{fig: query driven} demonstrates the effects of $\mathrm{DynConv}^{\mathtt{mask}}$ qualitatively.

Last, we also find that for query based approaches, HTC mask flow cannot bring further improvement on top of parallel $\mathrm{DynConv}^{\mathtt{mask}}$ architecture ($6$-th row).
This indicates that the proposed parallel $\mathrm{DynConv}^{\mathtt{mask}}$ enables adequate mask information flow propagating across queries in different stage for high quality mask generation. 
Therefore establishing explicit mask feature flow as HTC is redundant and is harmful for model efficiency.
In consideration of the speed-accuracy trade-off, we choose Fig.~\ref{fig: QueryInst} (c) as as the default instantiation of our QueryInst.

\section{Conclusion}

In this paper, we propose an efficient query based end-to-end instance segmentation framework, QueryInst, driven by parallel supervision on dynamic mask heads.
To our knowledge, QueryInst is the first query based instance segmentation method that outperforms previous state-of-the-art non-query based instance segmentation approaches. 
Extensive study proves that parallel mask supervision can bring great performance improvement without any decent for inference speed, while dynamic mask head with both shared query and MSA joints two sub-tasks of detection and segmentation naturally.
We hope this work can strength the understanding of query based frameworks and facilitate future research.

\begin{table}[t!]
    \centering
    \setlength{\tabcolsep}{4.2 pt}
    \begin{tabular}{c|c|cccc}
    \hline
    
    \hline

    \hline
    
    \rowcolor{mygray}
         Metrics
         & Aug. 
         & $3 \times$ 
         & $4 \times$ 
         & $5 \times$ 
         & $6 \times$ 
         \\
         \hline
         \hline
         \multirow{2}{*}{AP$^\mathtt{box}$} & $640 \sim 800$ 
         & $42.9$ & $42.5$ & $42.4$ & $42.7$ \\
         & $480 \sim 800$, w/ crop 
         & $42.7$ & & & $43.9$\\
         \hline
         \multirow{2}{*}{AP$^\mathtt{mask}$} & $640 \sim 800$ 
         & $38.6$ & $38.2$ & $38.1$ & $38.3$ \\
         & $480 \sim 800$, w/ crop 
         & $38.6$ & & & $39.4$\\
         \hline
         
         \hline
         
         \hline
    \end{tabular}
    \smallskip
    \caption{Object detection and instance segmentation performance of Mask R-CNN with ResNet-$101$-FPN backbone using training schedules from $2 \times$ ($180k$ iterations) to $6 \times$ ($540k$ iterations) and different data augmentations on COCO $\mathtt{val}$. The numbers under ``Aug." indicate the scale range of the shorter size of inputs with a stride of $32$.}
    \label{tab: diff schedule for mask rcnn.}
\end{table}

\begin{figure*}[t!]
    \centering
    \includegraphics[width=2.0\columnwidth]{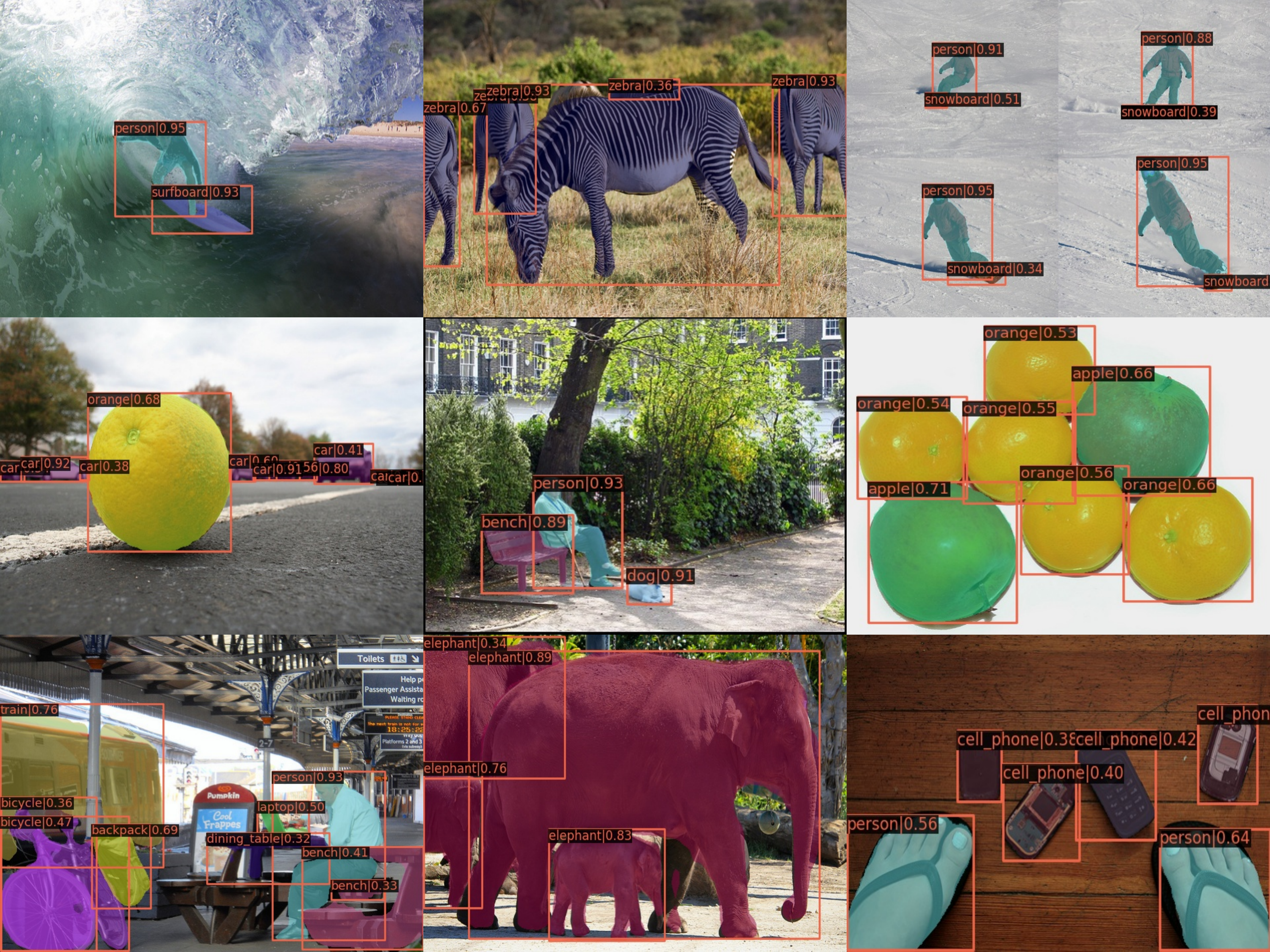}
    \caption{Object detection and instance segmentation qualitative results on COCO $\mathtt{val}$ split.}
    \label{fig: COCO}
\end{figure*}

\section*{Appendix}

\subsubsection*{Illustration of 4 Different Query and MSA \\ Configurations}

We study the impact of using shared query and shared MSA in the paper. 
Fig.~\ref{fig: share query and share msa.} gives an illustration for our ablation study. 
Fig.~\ref{fig: share query and share msa.} from left to right corresponds to configurations in Tab.~\ref{tab: queries and MSA ablation} in our paper from top to bottom.
Using the shared query and shared MSA configuration, QueryInst achieves the best performance in terms of both box AP and mask AP with the least additional overhead.

\subsubsection*{Training Time of QueryInst}

Here, we compare the training time of QueryInst with the state-of-the-art instance segmentation method HTC~\cite{HTC}. 
As shown in Tab~\ref{tab: training time}, under the same experimental configuration, QueryInst outperforms HTC using less training time.

\begin{figure*}[h]
    \centering
    \includegraphics[width=2.0\columnwidth]{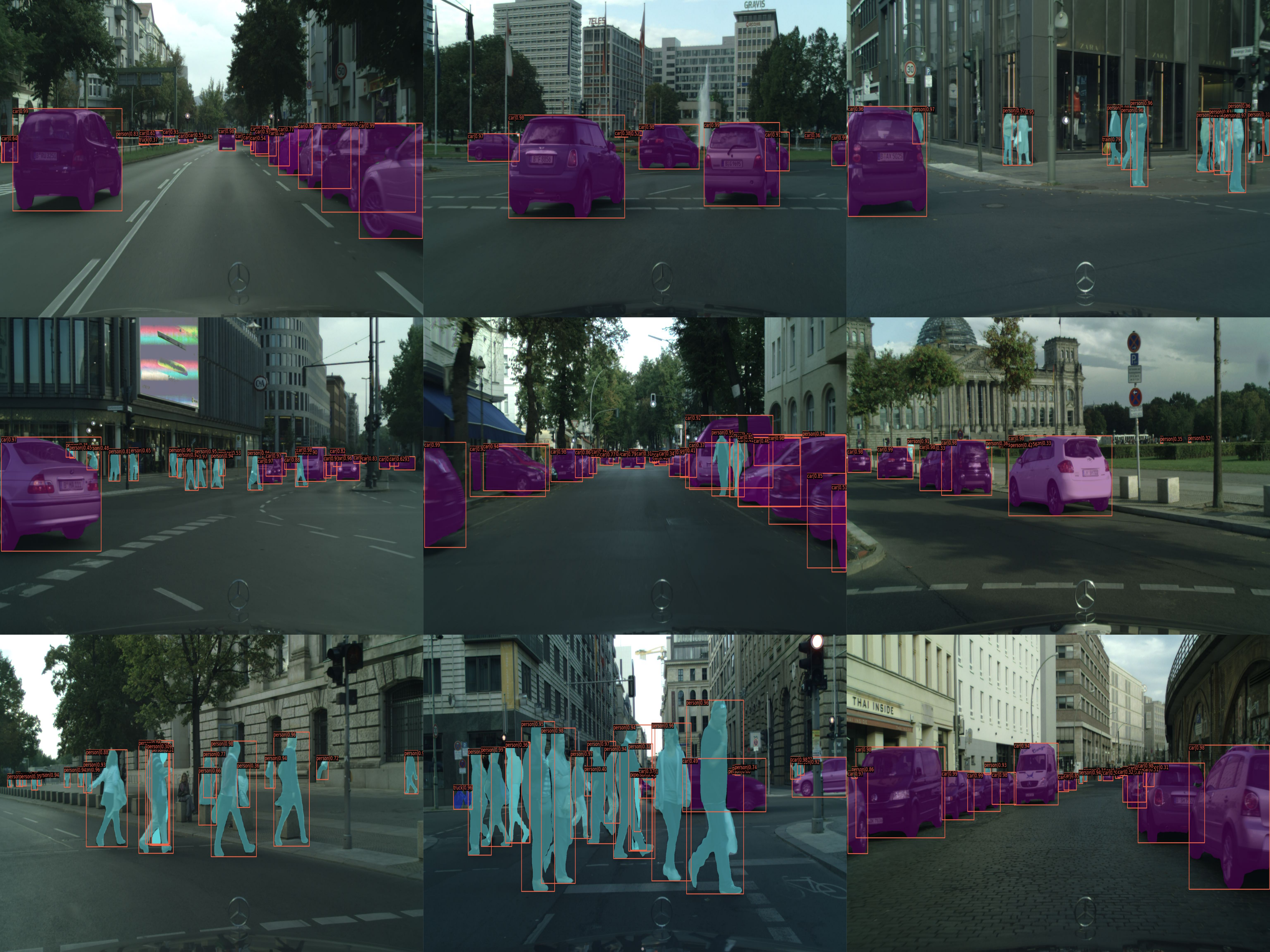}
    \caption{Object detection and instance segmentation qualitative results on Cityscapes $\mathtt{test}$ split.}
    \label{fig: Cityscapes}
\end{figure*}

\subsubsection*{Effects of Different Training Schedules and Data Augmentations to the Mask R-CNN Family}

In Tab.~\ref{tab: main results on test-dev}, we find that for Cascade Mask R-CNN and HTC, stronger data augmentation cannot bring significant improvements under the $3 \times$ training schedule. 
Here we give a detailed experimental study in Tab.~\ref{tab: diff schedule for mask rcnn.}.

We choose Mask R-CNN with ResNet-$101$-FPN backbone as a representative.
Using modest data augmentation ($640 \sim 800$), the $3 \times$ schedule works well for the model to converge to near optimum, the performance begins to degenerate under longer training schedules. 
These findings are consistent with \cite{RethinkingPretrain}.

Using stronger data augmentation ($480 \sim 800$, w/ crop) cannot bring further improvements under the $3 \times$ schedule, but can boost the performance as the training time becomes longer.

Compared with the Mask R-CNN family (Mask R-CNN, Cascade Mask R-CNN \& HTC), the proposed QueryInst can benefit from stronger data augmentation even under the $3 \times$ schedule. 
We will conduct further study of these properties in the future.

\subsubsection*{Qualitative Results on COCO}

	We provide some qualitative results on COCO~\cite{COCO} $\mathtt{val}$ split in Fig.~\ref{fig: COCO}.

\subsubsection*{Qualitative Results on Cityscapes}

	Fig.~\ref{fig: Cityscapes} gives some qualitative results on Cityscapes~\cite{Cityscapes} $\mathtt{test}$ split. 

\subsubsection*{Additional Visualization of Dynamic Mask Feature}

\begin{figure*}[t]
    \centering
    \includegraphics[width=2.0\columnwidth]{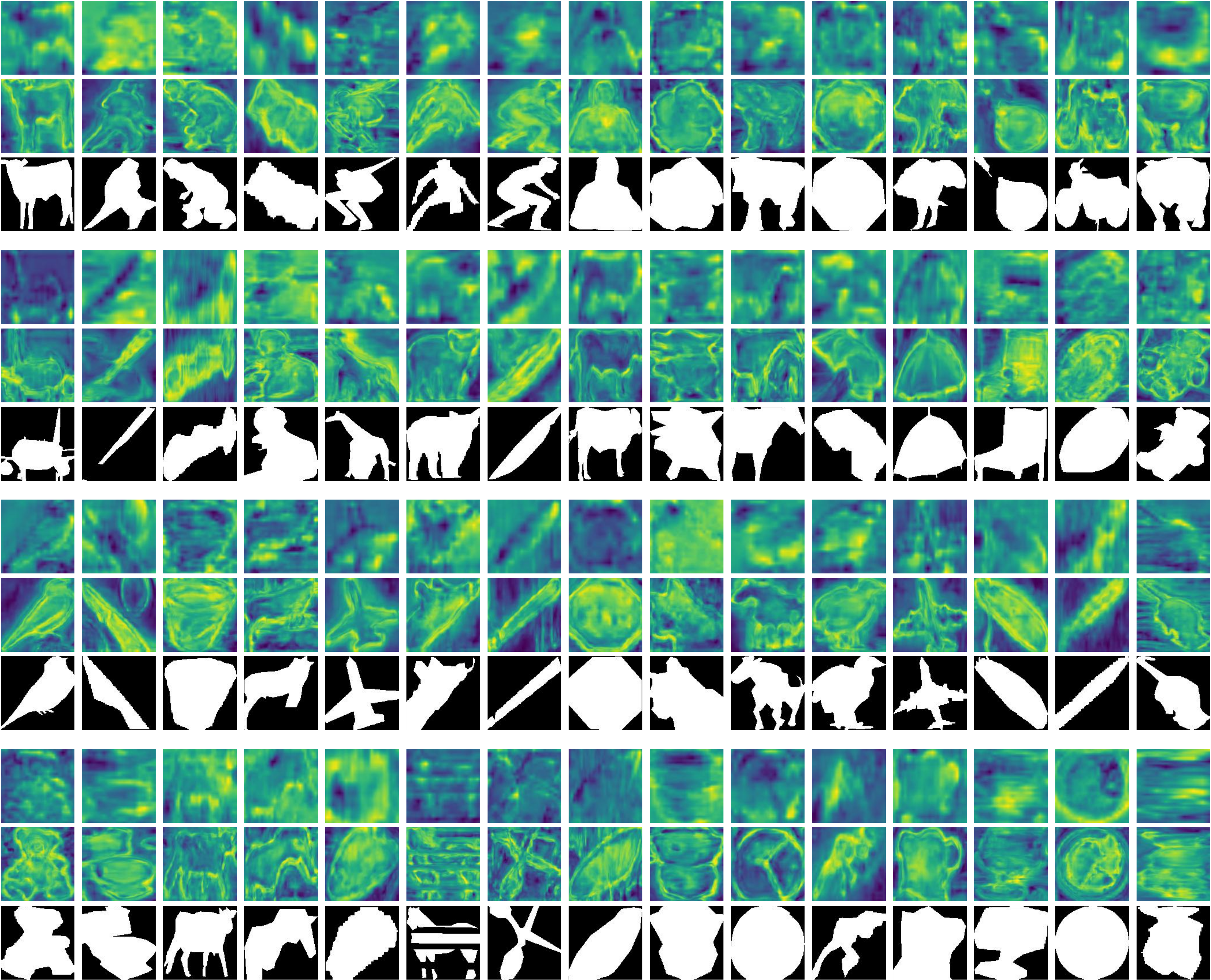}
    \caption{More visualization results on the study of $\mathrm{DynConv}^{\mathtt{mask}}$.}
    \label{fig: QueryVisualization}
\end{figure*}

In Fig.~\ref{fig: QueryVisualization}, we provide more visualization results to study the effects of $\mathrm{DynConv}^{\mathtt{mask}}$.
The first row shows mask features $\boldsymbol{x}^{\mathtt{mask}}$ directly extracted from FPN.
The second row shows mask features $\boldsymbol{x}^{\mathtt{mask*}}$ enhanced by queries in $\mathrm{DynConv}^{\mathtt{mask}}$. 
The last row is ground-truth instance masks.

\clearpage

{\small
\bibliographystyle{ieee_fullname}
\bibliography{QueryInst}
}

\end{document}